\documentclass[sigconf]{acmart}
\usepackage{booktabs}
\usepackage{multirow}
\usepackage{array}
\usepackage{enumitem}

\newcolumntype{P}[1]{>{\centering\arraybackslash}p{#1}}

\AtBeginDocument{%
  \providecommand\BibTeX{{%
    \normalfont B\kern-0.5em{\scshape i\kern-0.25em b}\kern-0.8em\TeX}}}

\setcopyright{acmlicensed}
\copyrightyear{2023}
\acmYear{2023}
\acmDOI{10.1145/3581783.3613784}

\acmConference[MM '23] {Proceedings of the 31st ACM International Conference on Multimedia}{October 29--November 3, 2023}{Ottawa, ON, Canada.}
\acmBooktitle{Proceedings of the 31st ACM International Conference on Multimedia (MM '23), October 29--November 3, 2023, Ottawa, ON, Canada}
\acmPrice{15.00}
\acmISBN{979-8-4007-0108-5/23/10}

\usepackage{amsmath}
\newcommand{\vect}[1]{\boldsymbol{#1}}

\def\eg{\textit{e.g.}}
\def\ie{\textit{i.e.}}

\def\etc{\textit{etc}}

\newcommand{\RS}[1]{{\color{black}{#1}}}

\begin{document}

\title{Efficient Labelling of Affective Video Datasets via Few-Shot \& Multi-Task Contrastive Learning}

%
\author{Ravikiran Parameshwara}
\email{ravikiran.parameshwara@canberra.edu.au}
\affiliation{%
  \institution{University of Canberra}
  \city{Canberra}
  \state{ACT}
  \country{Australia}
}
\author{Ibrahim Radwan}
\email{ibrahim.radwan@canberra.edu.au}
\affiliation{%
  \institution{University of Canberra}
  \city{Canberra}
  \state{ACT}
  \country{Australia}
}
\author{Akshay Asthana}
\email{akshay.asthana@seeingmachines.com}
\affiliation{%
  \institution{Seeing Machines Ltd.}
  \city{Canberra}
  \state{ACT}
  \country{Australia}
}
\author{Iman Abbasnejad}
\email{iman.abbasnejad@seeingmachines.com}
\affiliation{%
  \institution{Seeing Machines Ltd.}
  \city{Canberra}
  \state{ACT}
  \country{Australia}
}
\author{Ramanathan Subramanian}
\email{ramanathan.subramanian@ieee.org}
\affiliation{%
  \institution{University of Canberra}
  \city{Canberra}
  \state{ACT}
  \country{Australia}
}
\author{Roland Goecke}
\email{roland.goecke@ieee.org}
\affiliation{%
  \institution{University of Canberra}
  \city{Canberra}
  \state{ACT}
  \country{Australia}
}

\renewcommand{\shortauthors}{Parameshwara, et al.}


\begin{abstract}
Whilst deep learning techniques have achieved excellent emotion prediction, they still require large amounts of labelled training data, which are (a) onerous and tedious to compile, and (b) prone to errors and biases. We propose Multi-Task Contrastive Learning for Affect Representation (\textbf{MT-CLAR}) for few-shot affect inference. MT-CLAR combines multi-task learning with a Siamese network trained via contrastive learning to infer from a pair of expressive facial images (a) the (dis)similarity between the facial expressions, and (b) the difference in valence and arousal levels of the two faces. We further extend the image-based MT-CLAR framework for automated video labelling where, given one or a few labelled video frames (termed \textit{support-set}), MT-CLAR labels the remainder of the video for valence and arousal.  Experiments are performed on the AFEW-VA dataset with multiple support-set configurations; moreover, supervised learning on representations learnt via MT-CLAR are used for valence, arousal and categorical emotion prediction on the AffectNet and AFEW-VA datasets. The results show that valence and arousal predictions via MT-CLAR are very comparable to the state-of-the-art (SOTA), and we significantly outperform SOTA with a support-set $\approx$6\% the size of the video dataset.
\end{abstract}





\maketitle

%
%

\begin{figure}[h]
  \centering
  \includegraphics[width=0.8\linewidth]{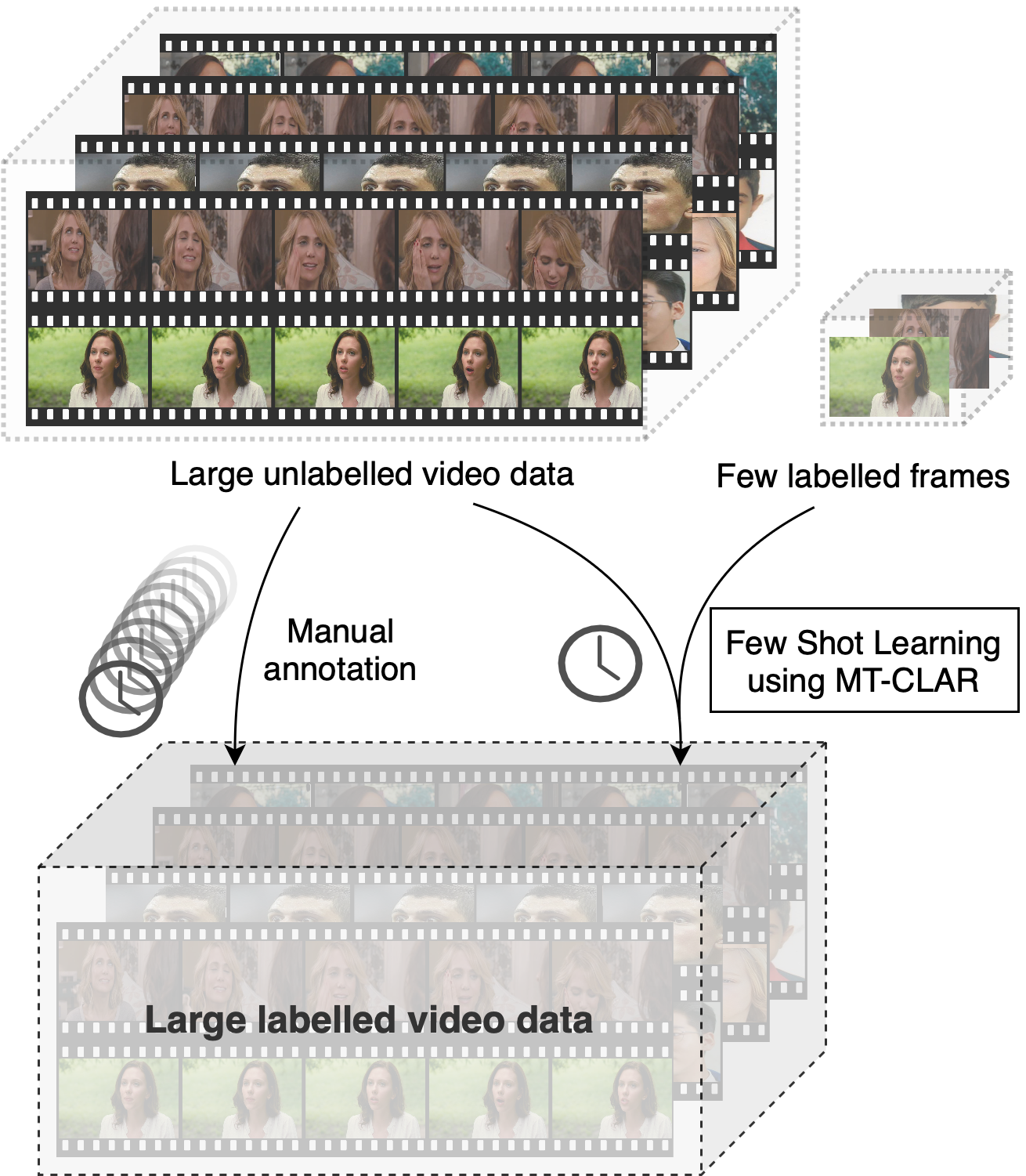} 
  \caption{Few-shot Affective Labelling: Annotating a large unlabelled video dataset is time-consuming and tedious. With MT-CLAR few-shot learning, utilising as few as $6\%$ labelled frames from the video dataset, we achieve excellent valence and arousal labelling for the remaining frames.}\vspace{-4mm}
  \label{fig:overview}
\end{figure}

\section{Introduction} 
\label{sec::introduction}

Automatically inferring human emotions is a challenging problem. Multiple modalities such as facial expressions~\cite{toisoul2021estimation}, speech~\cite{abbaschian2021deep}, and neural signals \cite{Bilalpur17,Shukla17,song2018eeg,Pandey2022} have been employed to this end. Emotion inference approaches typically use either \textit{categorical} (\ie, emotion classes) or \textit{dimensional} (\ie, the \emph{valence} and \textit{arousal} attributes) models; the latter captures subtle emotional variations on a continuous scale and is more flexible than the former. Advancements in emotion inference over the past two decades~\cite{li2020deep, tellamekala2019temporally,parameshwara2023examining} have enabled a transition from recognising emotions on \emph{acted} datasets involving manipulated emotional experiences, to \emph{in-the-wild} datasets capturing naturalistic or real-world settings.

The advent of deep learning approaches~\cite{toisoul2021estimation,narayana2022improve, narayana2023weakly} has significantly improved affect recognition performance. Nevertheless, deep learning algorithms require extensive labelled training data. Data labelling is a time-consuming, error-prone, costly and onerous task requiring skilled annotators to carefully scrutinise each sample. Annotations could provide a subjective judgement~\cite{gendron2018universality} of the presented emotion creating plausible bias. Moreover, \emph{evaluator lag} is a common problem in dynamic emotion annotation~\cite{huang2015investigation}, such that a temporal shift is found to better align the stimulus with the annotations in~\cite{tellamekala2019temporally, kollias2019deep}. Cumulatively, these issues can hinder models' efficacy to learn generalisable representations.


To address these challenges, we propose to use \textit{\textbf{Few-Shot Learning}} (FSL) as an alternative (Fig.~\ref{fig:overview}), which compensates for the shortage of annotated samples in the target domain~\cite{wang2020generalizing}. FSL algorithms learn from a few labelled examples and can generalise to new tasks with limited or no additional data. A \emph{support-set} comprising a few labelled samples per class is used to train the model to label \emph{query} (test) samples. 

\begin{figure*}[t]
  \centering
  \includegraphics[width=0.8\textwidth]{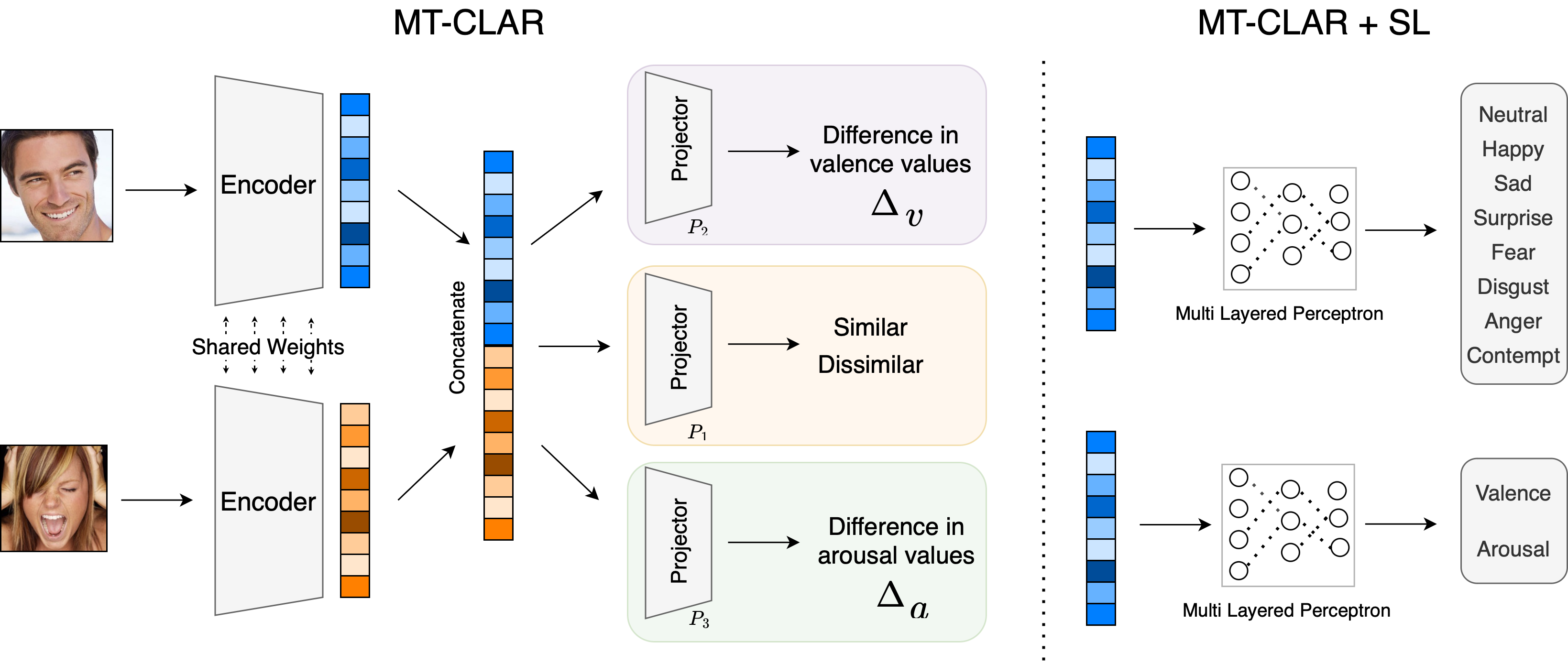} \vspace{-2mm}
  \caption{MT-CLAR overview: (Left) Differential estimation with MT-CLAR -- A pair of expressive facial images is passed through a Siamese network, and their embeddings are concatenated to estimate (1) whether the expressions are similar/dissimilar, (2) the valence differential ($\Delta_v$), and (3) the arousal differential ($\Delta_a$) between expressions. Learned representations are utilised for supervised learning from individual images (MT-CLAR + SL). (Right) MT-CLAR + SL: Either image embedding is fed to a Multi-Layer Perceptron to infer the emotion class, and estimate the valence and arousal values.} \vspace{-2mm}
  \label{fig:mt-clar_arch}
\end{figure*}

This study performs FSL via Multi-Task Contrastive Learning for Affect Representation (MT-CLAR), a novel approach to infer dynamic valence (\emph{val}, \ie, the extent of pleasure or sorrow induced by an emotional display) and arousal (\emph{asl}, \ie, the degree of physiological activation induced by the display) in videos. MT-CLAR involves a Siamese network trained via contrastive loss (Fig.~\ref{fig:mt-clar_arch} (left)), which captures the underlying (dis)similarity in a pair of expressive facial images. Using metric learning, MT-CLAR effectively learns intra-class similarities and inter-class differences. Leveraging multi-task learning, MT-CLAR primarily infers expressive facial pair similarity/dissimilarity in terms of categorical emotions, and secondarily predicts differentials in valence ($\Delta_v$) and arousal~($\Delta_a$). Utilising a few labelled \textit{anchor} video frames, and the estimated $\Delta_v$ and $\Delta_a$ from MT-CLAR, the remainder of a video can be automatically labelled for valence and arousal, respectively (Fig.~\ref{fig:fsl_overview}). 
MT-CLAR can also be integrated with supervised learning (MT-CLAR + SL), to predict categorical and dimensional labels for singleton images (Fig.~\ref{fig:mt-clar_arch} (right)). Overall, we make the following research contributions: 

\begin{enumerate}
    \item To the best of our knowledge, this study is the first to employ FSL-based approach to dynamic facial valence and arousal labelling in videos.
    Experiments on AffectNet \cite{mollahosseini2017affectnet} and AFEW-VA \cite{kossaifi2017afew} confirm that MT-CLAR generalises well, and can outperform the state-of-the-art with a {support-set} of only 6\% the size of a video dataset. 
    \item Different from the state-of-the-art~\cite{toisoul2021estimation,kossaifi2020factorised}, MT-CLAR employs a Siamese network trained via image pairs and contrastive loss to estimate a) emotional (dis)similarity, and b) valence ($\Delta_v$) and arousal~($\Delta_a$) differentials for an image pair. 
    \item \begin{sloppypar} MT-CLAR is further extended via supervised learning (MT-CLAR + SL) to 
    deduce categorical and dimensional emotion labels for singleton images as in~\cite{toisoul2021estimation,kossaifi2020factorised}. Extensive experiments confirm that MT-CLAR + SL achieves state-of-the-art results on multiple metrics for the AFEW-VA dataset~\cite{kossaifi2017afew} and highly competitive results on AffectNet~\cite{mollahosseini2017affectnet}.\end{sloppypar}
\end{enumerate}


%
%

\section{Related work}
\label{sec::related_work}

This section reviews literature relating to the use of metric learning, multi-task and few-shot learning for affect inference, to highlight the novelty of MT-CLAR. 

\subsection{Contrastive \& Metric Learning Approaches}
Contrastive learning involves weak supervision, where a model enforces like samples to be closer and unlike samples to be farther in the feature space. Contrastive loss was first proposed in~\cite{chopra2005learning} to approximate the semantic distance between a pair of facial images. Facial emotion contrast in the \emph{val}-\emph{asl} space was analysed in~\cite{kim2022emotion}. Using temporal sampling-based augmentation,~\cite{roy2021spatiotemporal} employed contrastive learning for facial expression recognition in videos. Contrastive learning has also been applied for speech emotion inference~\cite{lian2018speech}, cross-subject emotion inference from EEG signals~\cite{shen2022contrastive}, and to learn discriminative facial Action Unit representations~\cite{sun2021emotion}. 

Metric learning aims at learning a distance-based embedding, such that the relative distance between classes are preserved \cite{hilliard2018few}. Contrastive learning denotes a specific type of metric learning. In~\cite{liu2020saanet}, a metric learning framework was developed via a Siamese Network (SN) to examine fine-grained facial expression distinctions. A deep SN, capturing the local structure of an embedding, was employed for facial expression recognition in~\cite{hayale2021deep}. An SN, which incorporates latent facial attributes and long-term dynamics, was utilised for dimensional emotion prediction in~\cite{wang2017ast}.
 
\subsection{Multi-task Learning}
Multi-task learning (MTL) exploits task relatedness to learn shared representations applicable to multiple tasks. Individual task performance has been shown to improve via this shared representation~\cite{caruana1993multitask}. In~\cite{xia2015multi}, the secondary tasks of \emph{val} and \emph{asl} prediction is integrated into the primary task of categorical emotion inference. Dependencies among the tasks of \emph{val} and \emph{asl} prediction, emotion classification and Facial Action Unit detection were explored via MTL in~\cite{zhang2023multi}. In~\cite{chen2017multimodal}, MTL was used for multimodal \emph{val} and \emph{asl} prediction, and MTL performance was found to be superior than single-task counterparts. Superior results were observed in~\cite{jeong2022multi}, when a multimodal MTL model was employed for the three tasks of \emph{val}-\emph{asl} prediction, facial expression classification and Action Unit detection. 

\subsection{Few-Shot Learning for Affect Inference}
Few-shot learning relieves the burden of compiling large-scale annotated data~\cite{wang2020generalizing}. It aims to classify samples from the target domain using only a few labelled examples. While FSL has been widely applied to gesture recognition~\cite{pfister2014domain}, person identification~\cite{wu2018exploit}, video action recognition~\cite{careaga2019metric} \etc., attention to FSL-based emotion inference has increased only recently. {Meta learning} is an FSL technique, where a model learns generic inter-task information to adapt to new tasks with only a few samples~\cite{hochreiter2001learning}. FSL efficacy for facial expression recognition via meta learning was demonstrated in~\cite{ciubotaru2019revisiting}. An effective cross-domain FSL method is proposed in~\cite{zou2022facial}, where a two-stage learning framework is employed to infer compound facial expressions. Metric-based FSL was used to infer categorical emotions in scripted speech data~\cite{feng2021few, ahn2021cross}. FSL was applied to fine-grained \emph{val} and \emph{asl} prediction using physiological signals in~\cite{zhang2022few}. 

\subsection{MT-CLAR Novelty}
In contrast to prior work, the novel aspects of our study are as follows. (a) We synthesise via MT-CLAR robust affective representations characterising both singleton and pairwise facial images by combining multi-task and contrastive learning, (b) for the first time, estimation of pairwise similarity, valence and arousal differentials are incorporated as tasks additional to image-based emotion prediction achieved in~\cite{toisoul2021estimation,kossaifi2020factorised}, and (c) we employ FSL to generate affective labels for videos utilising a small support-set, and achieve SOTA predictions with as little as 6\% labelled frames.

%
%

\label{subsec::mt_clar}
\section{Proposed Framework}
\label{sec::proposed_framework}
This section describes MT-CLAR as depicted in Fig.~\ref{fig:mt-clar_arch} (left)), MT-CLAR + SL (Fig.~\ref{fig:mt-clar_arch} (right)), and the proposed FSL approach (Fig.~\ref{fig:fsl_overview}).


%
%
\subsection{Multi-Task Contrastive Learning}
\label{subsec::MTCL}
Contrastive learning aims to learn data representations that can discriminate similar vs.\ dissimilar samples; the objective is to pool similar samples together, while pushing dissimilar samples apart. It is known to yield high-quality representations for further processing~\cite{jing2020self, chen2020simple}. MT-CLAR employs SN to generate embeddings for a pair of expressive facial images. The SN comprises two identical sub-networks to compare the inputs. As shown in Fig.~\ref{fig:mt-clar_arch} (left), each sub-network comprises an \emph{encoder} for transforming the input image into a low-dimensional embedding. The two embeddings are concatenated and a \emph{projector} comprising linear layers is used to label the input pair as \textit{similar} or \textit{dissimilar}. Additionally, leveraging the efficiency and faster learning capabilities of MTL, the concatenated features are used for predicting \emph{val} and \emph{asl} differentials ($\Delta_v$, $\Delta_a$) for the input pair, through two corresponding \emph{projectors}.

%
%
\emph{\textbf{Encoder.}}
MT-CLAR employs EmoFAN \cite{toisoul2021estimation}, built on top of the Face Alignment Network (FAN) \cite{bulat2017far}, as the encoder, $Enc(\cdot)$. The input pair of images, $\vect{x_1}$ and $\vect{x_2}$, is mapped to the third-last layer of EmoFAN yielding representation vectors, $\vect{r_1} = Enc_{1}(\vect{x_1})$, and $\vect{r_2} = Enc_{2}(\vect{x_2})$, where $\vect{r_1}, \vect{r_2} \in \mathcal{R}^{256}$. As in a typical SN, $Enc_{1}(\cdot)$ and $Enc_{2}(\cdot)$ in the two streams share parameters and weights to produce two embeddings corresponding to the input images.

%
%
\emph{\textbf{Projector Network.}}
The vectors $\vect{r_1}$ and $\vect{r_2}$ obtained from $Enc_{1}(\cdot)$ and $Enc_{2}(\cdot)$, respectively, are concatenated to obtain $\vect{u} = \vect{r_1} \mathbin\Vert \vect{r_2}$, where $\vect{u} \in \mathcal{R}^{512}$. To classify the input images as (dis)similar and to predict $\Delta_v$ and $\Delta_a$, $\vect{u}$ is fed to three branched projector networks, $P_1(\cdot), P_2(\cdot),$ and $P_3(\cdot)$, which map $\vect{u}$ to three vectors $\vect{w_1} = P_1(\vect{u}), \vect{w_2} = P_2(\vect{u})$ and $\vect{w_3} = P_3(\vect{u})$, respectively. $P_1, P_2,$ and $P_3$ are Multi-Layer Perceptrons (MLPs), with four identical fully-connected (fc) layers, but a different number of neurons (2048, 1024, 512, and 128 neurons, respectively) in the last layer. MLPs $P_2$ and $P_3$ have a single terminal neuron to predict $\Delta_v$ and $\Delta_a$, while $P_1$ has two neurons to label (dis)similarity. The fc layer inputs are $z$-normalised and undergo ReLU activation.

%
%
\emph{\textbf{Loss Function.}}
Contrastive loss pulls intra-class embeddings closer and pushes inter-class embeddings apart~\cite{chen2020simple}. To learn representations of the input images $\vect{x_1}$ and $\vect{x_2}$, we apply contrastive loss, $\mathcal{L}_{cont}$ on vectors $\vect{r_1}, \vect{r_2}$ as follows: 
\vspace{-1mm}
\begin{equation}
\label{eq::loss_contrastive}
\mathcal{L}_{cont} = \frac{1}{N} \sum_{i=1}^N y_{i}(1 - d_{i}) + (1 - y_{i}) \max(0, d_{i} - m)
\vspace{-1mm}
\end{equation}
where $N, d_i$ and $m$, respectively, denote the batch size, cosine distance between $\vect{r_1},\vect{r_2}$, and the margin. $y_{i} = 0$ for dissimilar samples, and $y_{i} = 1$ for similar samples. Further, to classify the input pair as similar or dissimilar, we use cross-entropy loss ($\mathcal{L}_{ce}$) in $P_1$. 

As predicting $\Delta_v$ or $\Delta_a$ is a regression problem, the aim is to reduce the mean squared error (MSE), while simultaneously maximising the correlation between ground-truth and the predicted values. We propose maximising Concordance Correlation Coefficient (CCC), widely used in dimensional affect inference~\cite{mollahosseini2017affectnet, kossaifi2020factorised, toisoul2021estimation} to this end. To predict $\Delta_v$ and $\Delta_a$ in $P_2$ and $P_3$, respectively, we use a dynamically weighted loss function $\mathcal{L}_{\Delta}$, as proposed in \cite{parameshwara2023examining}. $\mathcal{L}_{\Delta}$ is a dynamically weighted sum of $\mathcal{L}_{mse}$ (squared $L2$ norm) and $\mathcal{L}_{ccc}$ (1 - CCC), and is given by: 
\begin{equation}
\label{eq::dyn_loss}
\mathcal{L}_{\Delta} = f \cdot \mathcal{L}_{mse} + g \cdot \mathcal{L}_{ccc}
\end{equation}
where $f$ and $g$ are dynamic weight functions given by:
\begin{equation}
\label{eq::loss_f_g}
    f = \alpha \left(\frac{i}{n}\right)^k \text{; } g = 1 - \left(\frac{i}{n}\right)^k
\end{equation}
where $i$ denotes the $i^{th}$ epoch within $n$ training epochs, and $\alpha \in \mathbb{R}$ and $k \in \mathbb{Z^+}$ are training hyper-parameters controlling the normalisation and non-linearity, respectively. Overall, to optimise MT-CLAR, we employ a cumulative loss function $\mathcal{L}$, defined as:
\vspace{-1mm}
\begin{equation}
\label{eq::loss_mt_clar}
    \mathcal{L} = \lambda_{1} \mathcal{L}_{cont} + \lambda_{2} \mathcal{L}_{ce} + \lambda_{3} \mathcal{L}_{\Delta_v} + \lambda_{4} \mathcal{L}_{\Delta_a}
\end{equation}
where $\mathcal{L}_{\Delta_v}$ and $\mathcal{L}_{\Delta_a}$ are the two $\mathcal{L}_{\Delta}$ corresponding to the $\Delta_v$ and $\Delta_a$ branches, respectively, and $\lambda_{i}, i \in \{1, 2, 3, 4\}$ are shake-shake regularisation coefficients \cite{gastaldi2017shake}, chosen randomly and uniformly in the range $[0, 1]$ at each training iteration. This ensures that the network does not prioritise any of the losses~\cite{toisoul2021estimation}. 

%
\begin{figure}[t]
  \centering
  \includegraphics[width=0.95\linewidth, height=4.4cm]{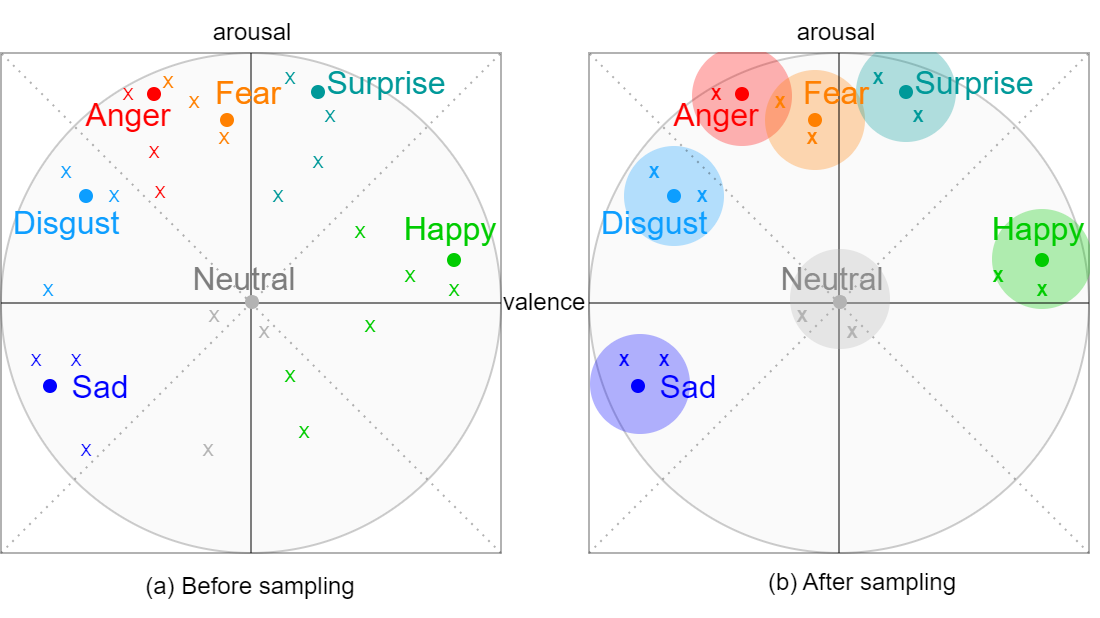} \vspace{-4mm}
 \caption{Mikel's wheel~\cite{mikels2005emotional}: Valence-arousal space visualisation (a) pre- and (b) post-sampling, where `x' denotes data points (hue: emotion category). Best viewed in colour.}  \label{fig:sampling}\vspace{-4mm}
\end{figure}

\emph{\textbf{Data Sampling.}}\label{Sec:DS}
Inspired by prior studies training SNs to recognise categorical emotions~\cite{hayale2019facial}, MT-CLAR takes as input an image pair with an associated emotion-based \emph{similar} or \emph{dissimilar} label. Given the  diverse and complex manifestation of emotions via facial expressions, it is crucial to identify relevant image pairs to train the SN. In the absence of representative data, SN performance is severely compromised~\cite{liu2020saanet}. 
Mikels' Wheel of Emotions~\cite{mikels2005emotional} is a visual representation of emotion classes in the valence-arousal $(V, A)$ space (Fig.~\ref{fig:sampling} (left)). Given an emotion category and its corresponding $(v, a) \in (V, A)$ in Mikel's wheel, a $d-$radius neighbourhood centred at $(v, a)$ is created (Fig.~\ref{fig:sampling} (right)). Data points within the neighbourhood are sampled for each emotion, while distant outliers are discarded, thus ensuring that only facial expression data points representative of every emotion category are considered. Upon sampling, facial image pairs with \emph{similar} or \emph{dissimilar} labels based on their emotion classes, are passed to MT-CLAR. 

%
%
\subsection{Few-Shot Learning}
\label{subsec::few_shot}

\begin{figure}[t]
  \centering
  \includegraphics[width=0.9\linewidth]{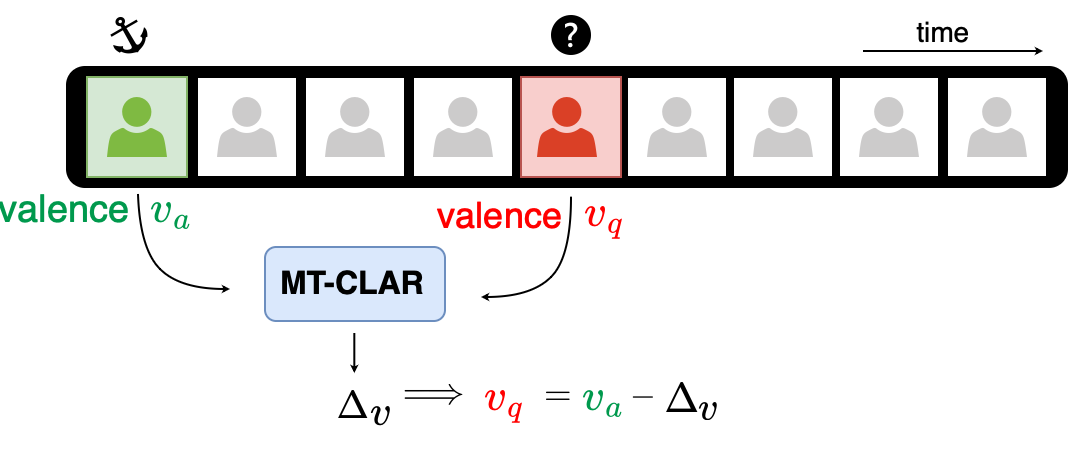} \vspace{-4mm}
  \caption{Few-shot learning: Given an \textit{anchor} video frame (green) with known \emph{val} label and a \textit{query} frame (red), MT-CLAR is used to predict the \emph{val} label of the query frame. An identical procedure is followed to predict the \emph{asl} label.} \vspace{-4mm}
  \label{fig:fsl_overview}
\end{figure}

Besides requiring large amounts of labelled data, traditional affect inference systems assume that a trained model can generalise well to the test set; however, this assumption is limiting, as conveyed by prior work on domain adaptation~\cite{domainadapt}. FSL methods present a promising alternative in this regard, as they can learn to generalise from limited samples known as the \textit{support-set}, obviating the need for large amount of labelled training data~\cite{tyukin2021demystification}. 

\begin{sloppypar}
In a metric-based FSL model, leveraging a \emph{support-set} $S = \{(\vect{x_1}, y_1), (\vect{x_2}, y_2), ..., (\vect{x_k}, y_k)\}$, comprising $k$ labelled samples, the goal is to predict label $\hat{y}_q$ for the query sample $\vect{x_q}$. A metric function $f(\vect{x};\theta)$ is defined to map the sample $\vect{x}$ to an embedding parameterised by $\theta$. The {distance} between $\vect{x_q}$ and each support sample $\vect{x_i} \in S$ is computed as $d(f(\vect{x_q}), f(\vect{x_i}))$. $\hat{y}_q$ corresponds to the closest support sample given by:
\end{sloppypar} 
\begin{equation}
    \hat{y}_q = \operatorname*{arg\,min}_i {d(f(\vect{x_q}), f(\vect{x_i}))}
\end{equation}

Our FSL methodology for dynamic \emph{val} and \emph{asl} labelling in videos is illustrated in Fig.~\ref{fig:fsl_overview}. We consider a support-set $S$ comprising video frames labelled for \emph{val} and \emph{asl}. Instead of computing the distance between $\vect{x_q}$ and each $\vect{x_i} \in S$, we only utilise an \emph{anchor set} $A_{S} \subset S$ for prediction. For a pair of frames, $\vect{x_i} \in A_S$ and $\vect{x_q}$, we obtain $\Delta_v$ via MT-CLAR defined as $\Delta_v = y_{i_{val}} - y_{q_{val}}$ where $y_{i_{val}}$ is the \emph{val} of $x_i$. Hence, $y_{q_{val}}$ (and likewise $y_{q_{asl}}$) are given by: 
\begin{align}
    y_{q_{val}} = y_{i_{val}} - \Delta_v \\
    y_{q_{asl}} = y_{i_{asl}} - \Delta_a
\end{align}
\vspace{-4mm}

    




\begin{figure}
  \centering
  \includegraphics[height=7.8cm,width=1.0\linewidth]{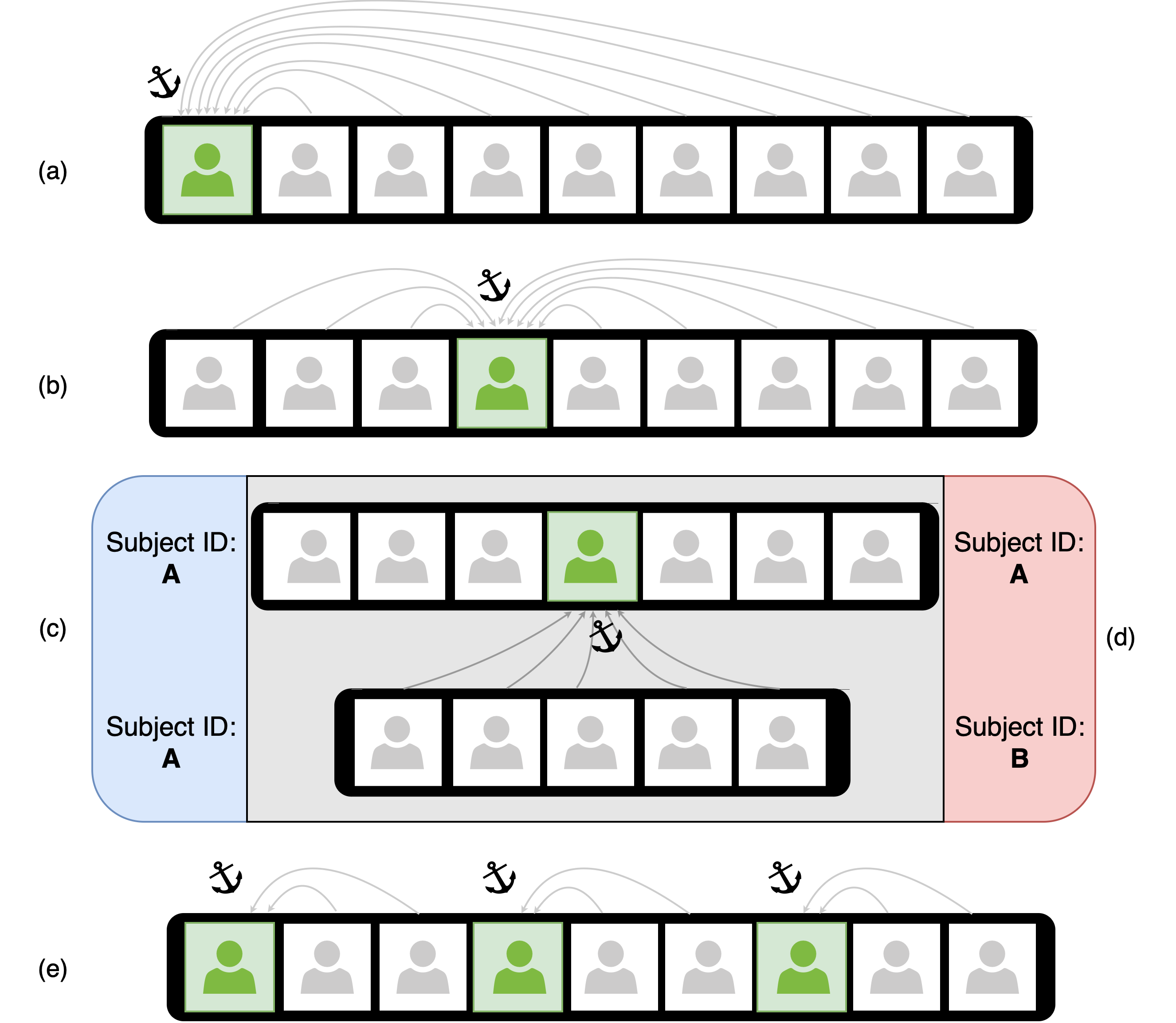} \vspace{-4mm}
  \caption{
  Possible anchor set $(A_{S})$ configurations with one or more anchor frames (green): (a) first video frame, (b) random video frame, random frame from a different video with (c) the same or (d) a different subject, and (e) recurring $n^{th}$ frame of a video.} \vspace{-3mm}
  \label{fig:fsl_configs}
\end{figure}

\paragraph{\textbf{Anchor Set Configurations}}
For dynamic emotion annotation, we only use an anchor set $A_{S} \subset S$ for precision and efficiency. Multiple $A_{S}$ configurations are shown in Fig.~\ref{fig:fsl_configs}, and described below:
\begin{itemize}[leftmargin=*]
    \item \textbf{First frame in video (Fig.~\ref{fig:fsl_configs} (a)):} For query frame $\vect{x_q}$, the first frame from the same video forms the anchor $\vect{x_a}$, or $A_S = \{\vect{x_a}$\}. 

    \item \textbf{Random frame in video (Fig.~\ref{fig:fsl_configs} (b)):} For query frame $\vect{x_q}$, $A_S = \{\vect{x_a}\}$, where $\vect{x_a}$ is a random frame from the same video. 


    \item \textbf{Random frame from subject-specific video (Fig.~\ref{fig:fsl_configs} (c)):} For query frame $\vect{x_q}, A_S = \{\vect{x_a}\}$, where $\vect{x_a}$ is a random frame from a (same or different) video with identical subject ID. 
    

    \item \textbf{Random frame of a different subject video (Fig.~\ref{fig:fsl_configs} (d)):} For query frame $\vect{x_q}, A_s = \{\vect{x_a}\}$, where $\vect{x_a}$ is a random frame from a video with different subject ID. 
    

    \item \textbf{Recurring $n^{th}$ frame in video (Fig.~\ref{fig:fsl_configs} (e)):} 
    Here, $A_S = \{\vect{x_{0}}, \vect{x_{n}}, \\  \vect{x_{2n}}, \cdots\}$. In this situation, the prediction of $\vect{x_q}$ may be based on immediately preceding $\vect{x_{kn}}$ (see Fig.~\ref{fig:fsl_configs} (e)), where $k = \lceil q/n \rceil$, or can be computed as the mean of the predictions obtained over all anchors in $A_S$.

    
        
        
\end{itemize}

%
%
\subsection{MT-CLAR + Supervised Learning}
\label{subsec::mt_clar_sl}
To evaluate the efficacy of the embeddings learned by the SN, and to enable MT-CLAR usage with singleton images, we combine MT-CLAR with supervised learning to synthesise the MT-CLAR + SL architecture. As seen in Fig.~\ref{fig:mt-clar_arch} (right), image features are passed through an MLP to predict its (a) (discrete) emotional class, and (b) (continuous) \emph{val}, \emph{asl} values.


\emph{\textbf{Architecture.}} The encoder embedding obtained for each image via the SN is input to MT-CLAR + SL. For classification, MT-CLAR + SL uses an MLP, $M_{C}(\cdot)$ to map input vector $\vect{x}$ to label vector $\hat{\vect{y_c}}$, such that $\hat{\vect{y_c}} = M_{C}(\vect{x})$, where $\vect{y_c} \in \mathcal{R}^8$, corresponding to the eight emotion classes. For predicting continuous \emph{val}, \emph{asl} values, MT-CLAR + SL uses another MLP, $M_{R}(\cdot)$, to map input vector $\vect{x}$ to vector $\vect{y_r}$, such that $\vect{y_r} = M_{R}(\vect{x})$, where $\vect{y_r} \in \mathcal{R}^ 2$, specifying \emph{val} and \emph{asl} estimates. Both $M_{C}(\vect{x})$ and $M_{R}(\vect{x})$ have four fc layers with 1024, 512, 256, and 128 neurons, respectively. Their inputs are $z$-normalised and ReLU activated before reaching the fc layers.

\emph{\textbf{Loss Function.}}
We apply cross-entropy loss, $\mathcal{L}_{CE}$ for classifying the input embedding into eight emotion classes. To infer continuous \emph{val}, \emph{asl} values, we apply a dynamically-weighted loss function $\mathcal{L}$, similar to Eq.~\ref{eq::dyn_loss}, given by:
\vspace{-1mm}
\begin{equation}
\label{eq::dyn_loss_supervised}
\mathcal{L}_{\Delta} = f \cdot \mathcal{L}_{mse} + g \cdot \mathcal{L}_{ccc}
\end{equation}
where $f$ and $g$ are dynamic weight functions as given by Eq.~\ref{eq::loss_f_g}. $\mathcal{L}_{mse}$ and $\mathcal{L}_{ccc}$ are given by:
\vspace{-2mm}
\begin{align}
    \mathcal{L}_{mse} = MSE_v + MSE_a  \\
    \mathcal{L}_{ccc} = 1 - \dfrac{CCC_v + CCC_a}{2}
\end{align}
where $MSE_v$ $(MSE_a)$ and $CCC_v$ ($CCC_a$) denote the mean square error and CCC, respectively, obtained with \emph{val} (\emph{asl}) prediction.


%
%

\section{Experimental Setup}
\label{sec:experimental_setup}

%
%
\subsection{Datasets}\label{Sec:Dataset}
The following datasets are used in this study.

%
%
\emph{\textbf{AffectNet}}~\cite{mollahosseini2017affectnet} is a large-scale, \emph{in-the-wild} facial expression dataset comprising 291,651 images annotated with both categorical labels, namely \emph{Neutral}, \emph{Happy}, \emph{Sad}, \emph{Surprise}, \emph{Fear}, \emph{Disgust}, \emph{Anger}, and \emph{Contempt}, and dimensional labels in terms of \emph{val}, \emph{asl} $\in [-1, 1]$ ratings. 
We employ AffectNet to train the MT-CLAR and MT-CLAR+SL models. Since the {AffectNet} test set is not released, we use the validation set of $\approx$ 4,000 images for evaluation.

\emph{\textbf{AFEW-VA}}~\cite{kossaifi2017afew}, a subset of AFEW \cite{dhall2012collecting}, is an affective video dataset with per-frame level \emph{val}, \emph{asl} annotations. It comprises 600 movie clips and $\approx$30,000 frames. The \emph{val}, \emph{asl} annotations are in the $[-10, 10]$ range, rescaled to $[-1, 1]$. We use AFEW-VA to annotate videos for \emph{val}, \emph{asl} values via FSL. It is also used to train the MT-CLAR+SL model for continuous emotion inference, using both subject-independent and subject-specific data-splits. The results for AFEW-VA are obtained via fine-tuning with using subject-independent and subject-specific 5-fold cross validation (5FCV). 

%
%
\subsection{Performance Metrics}

We employ multiple metrics for performance evaluation given the varied MT-CLAR outputs. To evaluate image (dis)similarity labelling and categorical emotion labelling performance, we use \textit{accuracy} (see Tables~\ref{tab:mt-clar},~\ref{tab:affectnet}). Further, as in~\cite{toisoul2021estimation, kossaifi2020factorised, kollias2020deep, mollahosseini2017affectnet, jang2019registration} to evaluate \emph{val} and \emph{asl} estimates, we use (a) Root Mean Square Error (RMSE), (b) Pearson Correlation Coefficient (PCC), (c) Concordance Correlation Coefficient (CCC), which incorporates PCC, but penalises correlated signals with different means, and (d) Sign Agreement (SAGR), a measure to evaluate if the sign of the predicted value matches with the target. Our aim is to minimise RMSE, while maximising PCC, CCC, and SAGR (see Tables \ref{tab:fsl_config}--\ref{tab:afewva}). 

If $\vect{y}$ and $\hat{\vect{y}}$ denote the ground truth and predicted labels, respectively, the above metrics are defined as: \vspace{-2mm}
%
\begin{eqnarray}
   RMSE(\vect{y}, \hat{\vect{y}}) & = & \sqrt{\mathbb{E}[(\vect{y} - \hat{\vect{y}})^2]} \\
    PCC(\vect{y}, \hat{\vect{y}}) & = & \frac{\mathbb{E}[(\vect{y} - \mu_{\vect{y}})(\hat{\vect{y}} - \mu_{\hat{\vect{y}}})]}{\sigma_{\vect{y}}\sigma_{\hat{\vect{y}}}} \\
    CCC(\vect{y}, \hat{\vect{y}}) & = & \frac{2\sigma_{\vect{y}}\sigma_{\hat{\vect{y}}}PCC(\vect{y}, \hat{\vect{y}})}{\sigma_{\vect{y}}^2 + \sigma_{\hat{\vect{y}}}^2 + (\mu_{\vect{y}} - \mu_{\hat{\vect{y}}})^2 \label{eq:cccloss}}
\end{eqnarray}
%
where $\mu_{\vect{y}}$ and $\sigma_{\vect{y}}$, respectively, denote the mean and the standard deviation of $\vect{y}$, and $\mathbb{E}$ denotes the expected value.

\subsection{Implementation Details}

We implement MT-CLAR using PyTorch \cite{paszke2019pytorch} software\footnote{URL for code repository: \url{https://github.com/ravikiranrao/MTCLAR-FSL}}. All models are trained using four Nvidia GeForce GTX 2080 Ti GPUs, each with 12GB RAM.  
The radius $d$ is set to 0.2 for generating the neighborhoods in Mikel's Wheel for the data sampling procedure. MT-CLAR inputs are resized to $288 \times 288$ pixels, and randomly cropped to 256 $\times$ 256 pixels. A random affine transformation is applied on the training images with a rotation of up to $\pm20$ degrees, translations up to $\pm20\%$ in both $x,y$ directions, scaling up to $\pm20\%$ and shearing up to $\pm10$ degrees. A horizontal flip is performed with 50\% chance. MT-CLAR is trained for 40 epochs with a batch size of 256 using the Adam~\cite{kingma2014adam} optimiser. The learning rate is scheduled based on plateau detection, with a base learning rate of 0.0001. It is decreased by a factor of 10 whenever a plateau is detected with the {patience} value set to 5. The margin $m$ used in contrastive loss (see Eq.~\ref{eq::loss_contrastive}) is set to 0.25. In the dynamic weight functions $f$ and $g$ (see Eq.~\ref{eq::loss_f_g}), the fine-tuned hyper-parameters are $k \in \{1, 2, 3\}$ and $\alpha \in \{1, 2, 20\}$. 

The input vector dimension for MT-CLAR + SL is 256. MT-CLAR + SL is trained for 60 epochs with a batch size of 512 using an Adam optimiser. The base learning rate is set to 0.001, and is decreased by a factor of 10 every 15 epochs. Hyperparameters $k,\alpha$ in the dynamic weight functions $f,g$ are fine-tuned identical to MT-CLAR.

%
%

\begin{table*}[!ht]
\fontsize{7}{9}\selectfont
\caption{Evaluating MT-CLAR design aspects via AffectNet. CE, Reg refer to cross entropy and regression loss, respectively.}
\label{tab:mt-clar}
\vspace{-2mm}
\begin{tabular}{@{}cllc@{}}
\toprule
\textbf{Data Sampler} & \textbf{Loss Function}           & \textbf{Task}                                                     & \textbf{Similarity Accuracy} \\ \midrule
No           & CE                     & Single (similarity)                                      & 0.53                    \\
No           & Contrastive             & Single (similarity)                                      & 0.60                    \\
No           & Contrastive + CE       & Single (similarity)                                      & 0.67                   \\
Yes          & Contrastive + CE       & Single (similarity)                                      & 0.69                    \\
Yes          & Contrastive + CE + Reg & Multi (similarity + $\Delta$ valence)                    & 0.70                    \\
Yes          & Contrastive + CE + Reg & Multi (similarity + $\Delta$ arousal)                    & 0.70                    \\
Yes          & Contrastive + CE + Reg & Multi (similarity + $\Delta$ valence + $\Delta$ arousal) & \textbf{0.72}                    \\ \bottomrule
\end{tabular}\vspace{3mm}
\centering
\caption{Few-shot affect inference on AFEW-VA with varying $S$ configurations. Best results in bold, while `*' denotes results significantly better than SOTA $(p < 0.05)$ as per Kolmogorov-Smirnov (KS) test.}\vspace{-2mm}
\label{tab:fsl_config}
\resizebox{\textwidth}{!}{%
\begin{tabular}{@{}clP{1.8cm}cccccccccc@{}}

\toprule
                      &                                                                      &              &          &                                     & \multicolumn{4}{c}{\textbf{Valence}} & \multicolumn{4}{c}{\textbf{Arousal}} \\ \cmidrule(l){6-13} 
\multirow{-2}{*}{\textbf{Row}} & \multirow{-2}{*}{\textbf{$A_S$ configuration}}   & \multirow{-2}{1.8cm}{ \centering  \textbf{$\vert S \vert$ (\% of total frames)}}                            & \multirow{-2}{*}{\textbf{Shot}} & \multirow{-2}{*}{ \centering \textbf{Finetuned}} & \textbf{RMSE $\downarrow$}      & \textbf{PCC $\uparrow$}        & \textbf{CCC $\uparrow$}        & \textbf{SAGR $\uparrow$}      & \textbf{RMSE $\downarrow$}      & \textbf{PCC $\uparrow$}        & \textbf{CCC $\uparrow$}        & \textbf{SAGR $\uparrow$}      \\ \midrule
0                                             &          SOTA (Toisoul et al.~\cite{toisoul2021estimation})                                                           &  -    &      -            & -                                                          & \textit{0.23 }       & \textit{0.70}        & \textit{0.69}        & \textit{0.65}       & \textit{0.22}       & \textit{0.67}        & \textit{0.66}        & \textit{0.81}       \\ \midrule
1                                             &                                                                      &      &                  & \textit{No}                                                          & \textit{\hspace{1.5mm}0.19 $^{*}$}       & \textit{0.68}        & \textit{0.68}        & \textit{0.59}       & \textit{0.21}       & \textit{0.66}        & \textit{0.64}        & \textit{0.78}       \\
2                                             & \multirow{-2}{*}{First frame of corresponding video}           &  \multirow{-2}{*}{91 (2.02\%)}    & \multirow{-2}{*}{One}  & Yes                                                         & 0.16       & 0.73        & 0.73        & 0.61       & 0.21       & 0.72        & 0.68        & 0.78       \\ \midrule
3                                             &                                                                      &      &                  & \textit{No}                                                          & \textit{\hspace{1.5mm}0.19 $^{*}$}       & \textit{\hspace{1.5mm}0.75 $^{*}$}        & \textit{\hspace{1.5mm}0.73 $^{*}$}        & \textit{0.63}       & \textit{0.22}       & \textit{0.64}        & \textit{0.64}        & \textit{0.83}       \\
4                                             & \multirow{-2}{*}{Random frame of corresponding video}           &  \multirow{-2}{*}{91 (2.02\%)}   & \multirow{-2}{*}{One}  & Yes                                                         & 0.14       & 0.83        & 0.82        & 0.63       & 0.19       & 0.74        & 0.72        & 0.84       \\ \midrule
5                                             &                                                                      &      &                  & \textit{No}                                                          & \textit{0.29}       & \textit{0.44}        & \textit{0.41}        & \textit{0.47}       & \textit{0.29}       & \textit{0.47}        & \textit{0.44}        & \textit{0.65}       \\
6                                            & \multirow{-2}{*}{Random frame from a subject-specific video}     &  \multirow{-2}{*}{34 (0.76\%)}  & \multirow{-2}{*}{One}  & Yes                                                         & 0.28       & 0.46        & 0.42        & 0.46       & 0.24       & 0.44        & 0.44        & 0.80      \\ \midrule
7                                             &                                                                      &     &                   & \textit{No}                                                          & \textit{0.56}       & \textit{0.01}        & \textit{0.00}           & \textit{0.11}       & \textit{0.36}       & \textit{0.01}        & \textit{0.01}        & \textit{0.82}       \\
8                                             & \multirow{-2}{*}{Random frame from a video of different subject} & \multirow{-2}{*}{34 (0.76\%)} & \multirow{-2}{*}{One}  & Yes                                                         & 0.59       & -0.14       & -0.03       & 0.14       & 0.33       & -0.01       & -0.01       & 0.82       \\ \midrule
9                                           &                                                                      &       &                 & \textit{No}                                                          & \textit{\hspace{1.5mm}0.19 $^{*}$}       & \textit{0.67}        & \textit{0.67}        & \textit{0.6}        & \textit{0.22}       & \textit{0.65}        & \textit{0.62}        & \textit{0.78}       \\
10                                           & \multirow{-2}{*}{Recurring $100^{th}$ frame of corresponding video}   &  \multirow{-2}{*}{96 (2.13\%)}   & \multirow{-2}{*}{Few}  & Yes                                                         & 0.15       & 0.77        & 0.76        & 0.6        & 0.21       & 0.72        & 0.67        & 0.78       \\ 
11                                            &                                                                      &        &                & \textit{No}                                                          & \textit{\hspace{1.5mm}0.19 $^{*}$}       & \textit{0.69}        & \textit{0.69}        & \textit{0.59}       & \textit{0.21}       & \textit{0.67}        & \textit{0.66}        & \textit{0.79}       \\
12                                            & \multirow{-2}{*}{Recurring $50^{th}$ frame of corresponding video}   & \multirow{-2}{*}{132 (2.93\%)}      & \multirow{-2}{*}{Few}  & Yes                                                         & 0.14       & 0.8         & 0.8         & 0.61       & 0.18       & 0.75        & 0.73        & 0.8        \\ 
13                                            &                                                                      &        &                & \textit{No}                                                          & \textit{\hspace{1.5mm}0.13 $^{*}$}       & \textit{\hspace{1.5mm}0.85 $^{*}$}        & \textit{\hspace{1.5mm}0.85 $^{*}$}        & \textit{0.66}       & \textit{\hspace{1.5mm}0.15 $^{*}$}       & \textit{\hspace{1.5mm}0.81 $^{*}$}        & \textit{\hspace{1.5mm}0.81 $^{*}$}        & \textit{\hspace{1.5mm}0.86 $^{*}$}       \\
14                                            & \multirow{-2}{*}{Recurring $20^{th}$ frame of corresponding video}      &  \multirow{-2}{*}{268 (5.96\%)} & \multirow{-2}{*}{Few}  & Yes                                                         &\textbf{ 0.09}       & \textbf{0.92}        & 0.91        & \textbf{0.66}       & \textbf{0.13}       & 0.86        & \textbf{0.86}        & 0.88       \\
15                                            &                                                                      &        &                & \textit{No}                                                          & \textit{\hspace{1.5mm}0.12 $^{*}$}       & \textit{\hspace{1.5mm}0.88 $^{*}$}        & \textit{\hspace{1.5mm}0.87 $^{*}$}        & \textit{0.64}       & \textit{\hspace{1.5mm}0.15 $^{*}$}       & \textit{\hspace{1.5mm}0.82 $^{*}$}        & \textit{\hspace{1.5mm}0.81 $^{*}$}        & \textit{\hspace{1.5mm}0.86 $^{*}$}       \\
16                                            & \multirow{-2}{*}{Recurring $10^{th}$ frame of corresponding video}      &  \multirow{-2}{*}{494 (10.98\%)} & \multirow{-2}{*}{Few}  & Yes                                                         &\textbf{ 0.09}       & \textbf{0.92}        & \textbf{0.92}        & 0.64      & \textbf{0.13}       & \textbf{0.87}        & \textbf{0.86}        & 0.87       \\ \midrule
17                                            &                                                                      &         &               & \textit{No}                                                          & \textit{\hspace{1.5mm}0.20 $^{*}$}        & \textit{\hspace{1.5mm}0.73 $^{*}$}        & \textit{0.71}        & \textit{0.62}       & \textit{\hspace{1.5mm}0.16 $^{*}$}       & \textit{\hspace{1.5mm}0.78 $^{*}$}        & \textit{\hspace{1.5mm}0.77 $^{*}$}        & \textit{\hspace{1.5mm}0.89 $^{*}$}       \\
18                                            & \multirow{-2}{*}{Recurring $10^{th}$ frame of corresponding video (mean)} & \multirow{-2}{*}{494 (10.98\%)}  & \multirow{-2}{*}{Few}  & Yes                                                         & 0.18       & 0.76        & 0.75        & 0.63       & 0.15       & 0.81        & 0.81        & \textbf{0.91}       \\ \bottomrule
\end{tabular}
}
\vspace{3mm}
\caption{Comparison with SOTA on {AffectNet}. Best results are in bold, and the second best are \underline{underlined}.}\vspace{-2mm}
\label{tab:affectnet}
\fontsize{7}{9}\selectfont
\begin{tabular}{@{}cccccccccc@{}}
\toprule
\multirow{2}{*}{\textbf{Model}} & \multirow{2}{*}{\textbf{Accuracy}} & \multicolumn{4}{c}{\textbf{Valence}} & \multicolumn{4}{c}{\textbf{Arousal}} \\ \cmidrule(l){3-10} 
                       &                           & \textbf{RMSE $\downarrow$}  & \textbf{PCC $\uparrow$} & \textbf{CCC $\uparrow$}  & \textbf{SAGR $\uparrow$} & \textbf{RMSE $\downarrow$}   & \textbf{PCC $\uparrow$} & \textbf{CCC $\uparrow$} & \textbf{SAGR $\uparrow$} \\ \cmidrule(r){1-10}
Mollahosseini \emph{et al.}\ (2019)~\cite{mollahosseini2017affectnet}                & 0.58                      & 0.37  & 0.66  & 0.60 & 0.74 & 0.41  & 0.54  & 0.34 & 0.65 \\
Jang \emph{et al.}\ (2019)~\cite{jang2019registration}              & -                         & 0.44  & 0.58  & 0.57 & 0.73 & 0.39  & 0.50  & 0.47 & 0.71 \\
Kollias \emph{et al.}\ (2020)~\cite{kollias2020deep}                    & \underline{0.60}                      & 0.37  & 0.66  & 0.62 & \underline{0.78} & 0.39  & 0.55  & 0.54 & 0.75 \\
Toisoul \emph{et al.}\ (2021)~\cite{toisoul2021estimation}               & \textbf{0.62}                      & \textbf{0.33}  & \textbf{0.73}  & \textbf{0.73} & \textbf{0.81} & \textbf{0.30}  & \textbf{0.65}  & \textbf{0.65} & \textbf{0.81} \\\midrule
\textbf{MT-CLAR + SL (Proposed)}                  &  0.56                     & \underline{0.36}   & \underline{0.67}   & \underline{0.67} & \underline{0.78} & \underline{0.32}  & \underline{0.60}  & \underline{0.60} & \textbf{0.81} \\
\bottomrule
\end{tabular}

\vspace{3mm}
\caption{Comparison with SOTA on {AFEW-VA} with 5FCV. \RS{Best results for each metric in the subject-independent condition are denoted in bold, and second-best \underline{underlined}. Best results in the subject-dependent condition are in bold.} }
\label{tab:afewva}\vspace{-2mm}
\fontsize{7}{9}\selectfont
\begin{tabular}{@{}llcccccccc@{}}
\toprule
\multirow{2}{*}{\textbf{Data-split strategy}} & \multirow{2}{*}{\textbf{Method}}        & \multicolumn{4}{c}{\textbf{Valence}}                                           & \multicolumn{4}{c}{\textbf{Arousal}}                                          \\ \cmidrule(l){3-10} 
                              &  & \textbf{RMSE $\downarrow$} & \textbf{PCC $\uparrow$} & \textbf{CCC $\uparrow$} & \textbf{SAGR $\uparrow$} & \textbf{RMSE $\downarrow$} & \textbf{PCC $\uparrow$} & \textbf{CCC$\uparrow$} & \textbf{SAGR $\uparrow$} \\ \midrule
\multirow{7}{*}{Subject-independent} & Kossaifi \emph{et al.}~(2017) \cite{kossaifi2017afew}         & 0.27              & 0.41           & -              & -               & 0.23              & 0.45           & -             & -               \\
 & Mitenkova \emph{et al.}~(2019) \cite{mitenkova2019valence}        & 0.40              & 0.33           & -              & -               & 0.41              & 0.42           & -             & -               \\
 & Handrich \emph{et al.}~(2020) \cite{handrich2020simultaneous}         & 0.28              & 0.58           & -              & -               & 0.26              & 0.46           & -             & -               \\
 & Kollias \emph{et al.}~(2020) \cite{kollias2020deep}         & 0.48              & 0.56           & -              & -               & 0.27              & 0.61           & -             & -               \\
 & Kossaifi \emph{et al.}~(2020) \cite{kossaifi2020factorised}         & 0.24              & 0.55           & \underline{0.55}           & \underline{0.64}            & 0.24              & 0.57           & \underline{0.52}          & 0.77            \\
 & Toisoul \emph{et al.}~(2021) \cite{toisoul2021estimation}           & \underline{0.23}              & \textbf{0.70}           & \textbf{0.69}           & \textbf{0.65}            & \underline{0.22}              & \textbf{0.67}           & \textbf{0.66}          & \textbf{0.81}            \\
\cmidrule(l){2-10}
 & \textbf{MT-CLAR + SL (Ours)}             &   \textbf{0.21}                &  \underline{0.69}              &  0.46              & 0.58                &  \textbf{0.19}                 & \underline{0.62}               & 0.42              & \underline{0.78}                \\ \midrule

\multirow{2}{*}{Subject-dependent} & Parameshwara \emph{et al.}~(2023) \cite{parameshwara2023examining} & 0.13              & 0.89           & 0.89          & -               & 0.12             & \textbf{0.93}           & \textbf{0.93}          & -               \\ 
& \textbf{MT-CLAR + SL (Ours)}        &  \textbf{0.12}             &  \textbf{0.90}          &  0.89             &  0.67              &  0.12             & 0.88           & 0.88             & 0.87              \\ \bottomrule
\end{tabular}

\end{table*}

\section{Results and Discussion}
\label{sec:results}

%
%
\subsection{Evaluating the MT-CLAR Design}
\label{subsec::results_mt_clar}

MT-CLAR compares facial image pairs (Fig.~\ref{fig:mt-clar_arch} (left)), and generates similarity labels and $\Delta_v, \Delta_a$ values. Table~\ref{tab:mt-clar} shows the impact of (a) optimising multiple loss functions, and (b) single vs.\ multi-task learning on similarity labelling performance on AffectNet.

\emph{\textbf{Loss Functions.}} 
Eq.~\ref{eq::loss_mt_clar} specifies that MT-CLAR is optimised by cumulatively minimising four losses. Projector $P_1$ employs cross-entropy loss $\mathcal{L}_{ce}$ to estimate (dis)similarity. However, an accuracy of only 0.53 is achieved on simply optimising for $\mathcal{L}_{ce}$. As the cross-entropy loss lacks robustness to noisy labels and poor margins~\cite{zhang2018generalized,liu2016large}, we additionally employ contrastive loss, $\mathcal{L}_{cont}$, which improves the accuracy by 7\%. Jointly minimising the two losses further improves accuracy by 7\%, while adding the sampling procedure (Sec.~\ref{subsec::MTCL}) results in a similarity accuracy of 0.69. 

\emph{\textbf{Single vs.\ Multi-task Learning.}}
In the above cases, MT-CLAR performs the solitary task of similarity labelling on AffectNet. Introducing the additional task of $\Delta_v$ or $\Delta_a$ prediction entails (1) feeding the ground-truth facial \emph{val} or \emph{asl} values to MT-CLAR, (2) integrating $\mathcal{L}_{cont}+\mathcal{L}_{ce}$ with either $\mathcal{L}_{\Delta_v}$ or $\mathcal{L}_{\Delta_a}$, and (3) leveraging the relationships between similarity and $\Delta_v$/$\Delta_a$ labelling. Consequently, an improvement in similarity accuracy of 1\% is noted in either case. Finally, combining the multiple tasks of similarity, ${\Delta_v}$ and ${\Delta_a}$ labelling entails optimising for the cumulative loss given by Eq.~\ref{eq::loss_mt_clar}, which achieves the best similarity accuracy of 0.72.

Overall, Table~\ref{tab:mt-clar} reveals that cumulatively minimising the considered losses plus exploiting the task-relatedness among similarity, \emph{val} and \emph{asl} differential estimation tasks benefit the primary MT-CLAR task of image similarity labelling.


%
%
\subsection{FSL-based Video Affect Labelling}
\label{subsec::results_fsl}

\begin{figure*}[t]
    \centering
    \includegraphics[width=\linewidth]{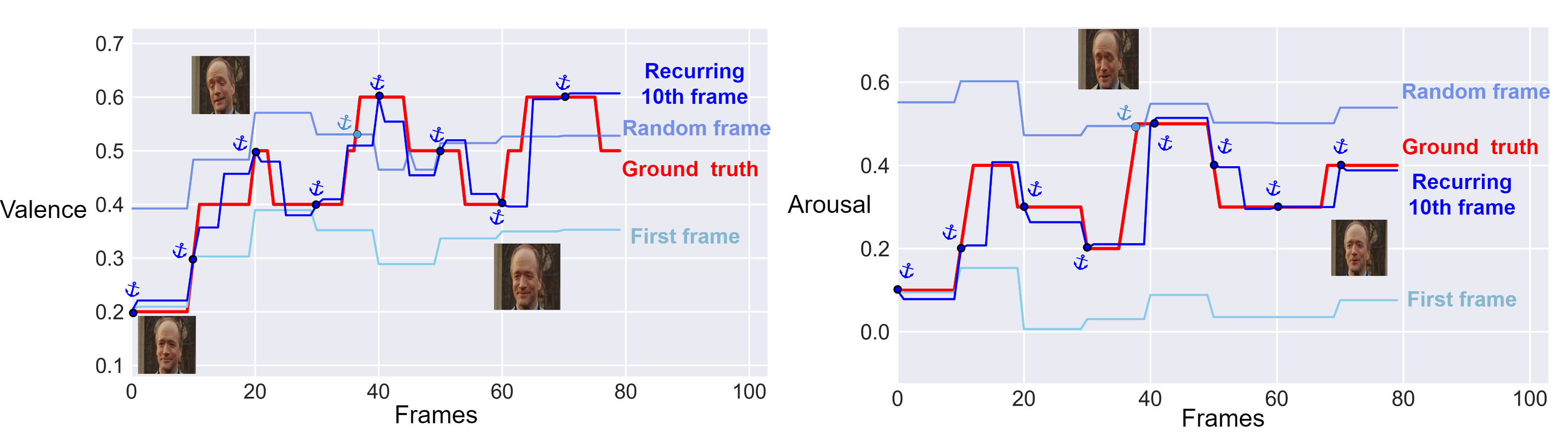}\vspace{-3mm}
    \caption{Continuous affect prediction in videos: Visualisation of the predicted \emph{val} (left) and \emph{asl} (right) values with multiple $A_S$ configurations, namely first frame, random frame, and recurring $10^{th}$ frame in an exemplar AFEW-VA video.}\vspace{-2mm}
    \label{fig:qualitative}
\end{figure*}

This study is the first work to generate \emph{val} and \emph{asl} labels for videos via FSL utilising a labelled \emph{anchor set} (Sec.~\ref{subsec::few_shot}). Whilst there are no competing methods to this end, Table~\ref{tab:fsl_config} nevertheless presents interesting insights regarding the impact of the anchor set configuration on the precision of the \emph{val} and \emph{asl} estimates. As per Sec.~\ref{Sec:Dataset}, all results in Table~\ref{tab:fsl_config} are obtained by training MT-CLAR on AffectNet, and evaluating the model on the AFEW-VA test set (mean values obtained over 5FCV). For each $A_{S}$ configuration, results are reported without and with finetuning on the AFEW-VA train set. Notably, the `No' rows correspond to conditions where only a specified number of AFEW-VA anchor frames (equal to $\vert S \vert$) are available. Results significantly better than SOTA~\cite{toisoul2021estimation} are denoted via a `*'.

We make the following remarks from Table~\ref{tab:fsl_config}. Focusing on the `No' rows, measures very comparable to SOTA are obtained including a significantly better RMSE (\emph{val}) when only the first video frame is employed as anchor. Using any random video frame as anchor further improves measures compared to SOTA with significantly better RMSE, PCC and CCC metrics generated; note that only 2\% of AFEW-VA are labelled in either case. Expectedly, poorer measures are noted in the limiting case when an anchor corresponding to the same subject ID is utilised as anchor, in which case $<1\%$ of labelled AFEW-VA frames are used. The lowest measures are observed in the extremely challenging case where an anchor corresponding to a different ID is employed, with \emph{asl}-related metrics faring better than \emph{val} metrics. Overall, these results are testimony to the assertion that FSL-based emotion inference \textit{in-the-wild} is highly difficult.  

Measures better than SOTA are obtained when multiple video frames are employed as anchors, significantly outcompeting SOTA for all-but-one measure with as few as 5.96\% labelled anchor frames. Comparing the results of rows 15 and 17, \ie, when using the most recent anchor vs.\ computing the mean predicted value across all anchors (recurring $10^{th}$ frames), respectively, the former approach is found to be optimal with respect to most measures.

The `Yes' rows in Table~\ref{tab:fsl_config} correspond to the condition where MT-CLAR is finetuned using AFEW-VA (4 out of 5 folds constituting the training set). Evidently, for all-but-one $A_S$ configurations, similar or better metrics as compared to no finetuning are obtained with MT-CLAR finetuning. The notable exception is when FSL is attempted with an anchor frame corresponding to a different subject ID (row 8); these results reveal that while model finetuning is in general beneficial, the choice of anchor(s) for few-shot learning is nevertheless critical, and can considerably impact prediction results. Furthermore, inadequate MT-CLAR labelling performance when the anchor frame corresponds to a different subject ID conveys that pairwise expressive face comparisons become easier when identity-related facial variations are accounted for, and identity-related facial representations, capturing global facial structure, are utilised by the MT-CLAR network to make predictions relating to emotions, which are characterised by the local facial structure. 

The best measures with MT-CLAR finetuning are obtained when the anchor set comprises recurring $20^{th}$ or $10^{th}$ frames in the video to be labelled. Almost identical measures for \emph{val} and \emph{asl} are obtained in either configuration, implying that close-to-peak affective labelling performance is achieved with MT-CLAR even as only $\approx$6\% frames in a video dataset are labelled.

\subsection{MT-CLAR + SL Prediction on Images}
\label{subsec::results_mt_clar_sl}
For MT-CLAR to be used with images as in~\cite{toisoul2021estimation,kossaifi2020factorised} and to validate the observation that contrastive learning generates robust, high-quality representations~\cite{jing2020self, chen2020simple}, we combine MT-CLAR with supervised learning in the MT-CLAR + SL architecture (Fig.~\ref{fig:mt-clar_arch}). When applied on images, it estimates their (a) emotion category labels, (b) \emph{val}, and (c) \emph{asl} labels. Results for MT-CLAR + SL and prior methods on the AffectNet dataset are presented in Table~\ref{tab:affectnet}.

For categorical emotion labelling, 
MT-CLAR's performance is comparable to other models. For \emph{val} estimation, MT-CLAR achieves the second-best performance w.r.t.\ four \emph{val} metrics, including an equal second-best SAGR as~\cite{toisoul2021estimation}. For \emph{asl}, we again achieve the second-best performance w.r.t.\ three metrics, and an equal-best SAGR as~\cite{toisoul2021estimation}. These results confirm that MT-CLAR + SL predictions are comparable to the state-of-the-art. 

Table~\ref{tab:afewva} presents continuous \emph{val} and \emph{asl} labelling results for MT-CLAR + SL on the AFEW-VA video dataset and comparisons with SOTA. We consider two 5FCV data-split strategies for the AFEW-VA dataset: subject-dependent and subject-independent. While both involve mutually exclusive training and test sets, the subject-independent setting also involves mutually exclusive subject IDs so as to preclude a \textit{data leak} from the training sets to the test set.

Observing Table~\ref{tab:afewva}, we make the following remarks. Consistent with Table~\ref{tab:mt-clar}, predictions in the subject-dependent setting are much better than those in the subject-independent setting. In comparison to other models, MT-CLAR + SL achieves the lowest \emph{val} RMSE and the second-best PCC. For \emph{asl}, we again obtain the lowest RMSE and the second-best PCC and SAGR. Considering subject-dependent splits, we outperform the image-based framework proposed in~\cite{parameshwara2023examining} w.r.t.\ \emph{val} RMSE and PCC, and obtain an identical CCC. 

Tables~\ref{tab:affectnet} and~\ref{tab:afewva} reveal that MT-CLAR + SL, designed to enable the MT-CLAR model to predict both categorical \emph{and} continuous emotion labels for singleton images, is highly competitive compared to other models exclusively designed to this end.




%
%
\subsection{Research Challenges \& Opportunities}
The empirical results confirm that MT-CLAR (a) enables accurate annotation of continuous \emph{val} and \emph{asl} values in videos when a labelled support-set is available (Table~\ref{tab:fsl_config}), and (b) achieves competitive \emph{val} and \emph{asl} level estimation for singleton images (Table~\ref{tab:affectnet}) and video frames (Table~\ref{tab:afewva}). Labelling images for emotion category, \emph{val}, and \emph{asl} enables automated emoji generation~\cite{AliWWHB17}, while dynamic affect labelling in videos greatly eliminates human effort and bias, and enables applications such as highlights detection~\cite{Qi2021}.  

Still, precisely estimating continuous \emph{val}, \emph{asl} levels from \textit{in-the-wild} videos presents a significant challenge, even if excellent RMSE, PCC, CCC and SAGR metrics are achieved over the test set (see Fig.~\ref{fig:qualitative}). The figure presents true and MT-CLAR-predicted \emph{val} (left) and \emph{asl} (right) values with multiple $A_S$ configurations. As per Table~\ref{tab:fsl_config}, \emph{val} and \emph{asl} estimates employing the first video frame as anchor already compare well with SOTA~\cite{toisoul2021estimation}; however, a considerable gap between true \emph{val} and \emph{asl} levels, and first frame-based MT-CLAR predictions can be noted for a majority of the considered video. 

Consistent with Table~\ref{tab:fsl_config} results, we note that the true \emph{val} and \emph{asl} trends are captured better with a random frame anchor. MT-CLAR predictions employing recurrent $10^{th}$ frames as anchor are far better than~\cite{toisoul2021estimation} from Table~\ref{tab:fsl_config}; whilst correspondingly, the dark blue curve best aligns with the (true) red curve for both \emph{val} and \emph{asl} prediction, the estimates are still far from precise. Overall, Fig.~\ref{fig:qualitative} clearly reveals that the RMSE, PCC, CCC and SAGR metrics are rather coarse-grained for the arduous problem of dynamic emotion inference, and even excellent results achieved w.r.t.\ these measures does not imply generation of precise estimates. Thus, worthy objectives for future work in this direction would be to (1) attempt precise affect predictions with few annotations, and (2) explore alternate performance metrics to better validate the precision of estimates. 

\section{Conclusion}\label{sec:conclusion}
Extensive empirical validation confirms that the  MT-CLAR framework achieves (1) state-of-the-art RMSE, PCC, CCC and SAGR metrics for dynamic emotion labelling in videos when a labelled support-set is available, and (2) competitive performance with respect to state-of-the-art for emotion class, \emph{val} and \emph{asl} estimation from single images/video frames. Nevertheless, qualitative examination reveals that further research is needed in the domain of continuous affect prediction, and alternate performance measures need to be explored for a rigorous performance evaluation. 

Future work will (a) extend MT-CLAR to include spatio-temporal information characterising videos, (b) evaluate MT-CLAR across video datasets to further verify its generalisability, and (c) use insights from this study to develop a {video annotation tool} that facilitates the affective labelling of large datasets with minimal human effort (\eg, an expert annotating a few \textit{keyframes} followed by the automated generation of affect labels for the remainder of the video).

\begin{acks}
This research is partially funded by the Australian Government through the Australian Research Council’s Discovery Projects funding scheme (project DP190101294).
\end{acks}

\bibliographystyle{ACM-Reference-Format}
\bibliography{ravikiran}




\end{document}